\documentclass{article}

\usepackage{arxiv}

\usepackage[utf8]{inputenc} 
\usepackage[T1]{fontenc}    
\usepackage{hyperref}       
\usepackage{url}            
\usepackage{booktabs}       
\usepackage{amsfonts}       
\usepackage{nicefrac}       
\usepackage{microtype}      
\usepackage{lipsum}
\usepackage[utf8]{inputenc} 
\usepackage[T1]{fontenc}    
\usepackage{hyperref}       
\usepackage{url}           
\usepackage{booktabs}      
\usepackage{amsfonts}     
\usepackage{nicefrac}       
\usepackage{microtype}     
\usepackage{lipsum}
\usepackage{tikz}
\usepackage[noend]{algpseudocode}
\usepackage{algorithmicx, algorithm}
\usepackage{xcolor}
\usepackage{soul, color}
\usepackage[misc]{ifsym}
\usepackage{graphicx}
\usepackage{epstopdf}
\usepackage{subfigure}
\usepackage{amsmath}
\usepackage{todonotes}
\usepackage{comment}
\usepackage{lineno}
\soulregister\ref7
\soulregister\cite7

\graphicspath{ {./Figures/} }

\title{A hybrid data driven-physics constrained Gaussian process regression framework with deep kernel for uncertainty quantification}

 \author{
 Cheng Chang \thanks{Department of Mathematics and Institute of Mathematical Sciences, The Chinese University of Hong Kong. Email: chengchang@link.cuhk.edu.hk. Address: Room G08, Lady Shaw Building, The Chinese University of Hong Kong, Shatin, N.T., Hong Kong},~~
  Tieyong Zeng \thanks{Department of Mathematics, The Chinese University of Hong Kong. Email: zeng@math.cuhk.edu.hk. Address: Room 225, Lady Shaw Building, The Chinese University of Hong Kong, Shatin, N.T., Hong Kong}
 }
\begin{document}
\maketitle

\begin{abstract}
Gaussian process regression (GPR) has been a well-known machine learning method for various applications such as uncertainty quantifications (UQ). However, GPR is inherently a data-driven method, which requires sufficiently large dataset. If appropriate physics constraints (e.g. expressed in partial differential equations) can be incorporated, the amount of data can be greatly reduced and the accuracy further improved. In this work, we propose a hybrid data driven-physics constrained Gaussian process regression framework. We encode the physics knowledge with Boltzmann-Gibbs distribution and derive our model through maximum likelihood (ML) approach. We apply deep kernel learning method. The proposed model learns from both data and physics constraints through the training of a deep neural network, which serves as part of the covariance function in GPR. The proposed model achieves good results in high-dimensional problem, and correctly propagate the uncertainty, with very limited labelled data provided. 
\end{abstract}


\section{Introduction}

In many practical fields like engineering, finance and physics, we need to explore the effects of uncertainties borne by input or parameters, which is the task of the uncertainty quantification (UQ). In a specific UQ problem, if the system is represented as stochastic partial differential equations, then stochastic Galerkin methods like generalised polynomial chaos (gPC) \cite{UQDL070} can be used to obtain statistics of the distribution of the unknown solution. As long as a numerical or analytical solver for the system is available, the renowned Monte Carlo (MC) method \cite{UQDL071} can be applied to obtain a distribution of the quantities of interest. However, that is usually prohibitively expensive due to the slow convergence of the Monte Carlo method, despite the fact that the running of the computer code for solving some complicated systems one single time might take days. Thus, to some extent we may be willing to compromise accuracy to reduce running time through formulating a cheap surrogate \cite{UQDL021}. That is where Gaussian process regression (GPR) comes in.

GPR has long been studied as a surrogate modelling tool in the UQ field. As a famous machine learning method, GPR has been well developed due to its various advantages, for example, it uses Bayesian inference and is able to provide an error bar instead a single point estimation, and it makes use of adaptive basis function, compared with fixed basis used in deep neural networks \cite{RW}. A variety of adaptions of GPR have been proposed to suit the needs in UQ applications. In \cite{UQDL021}, the authors divide the whole domain of interest into smaller elements to reduce the uncertainty in each smaller, local element, and provide procedures to calculate and combine statistics obtained from those elements. In \cite{UQDL014}, the authors make assumptions that the correlation function can be decomposed into product of those of input parameters, spatial and temporal variables, to drastically improve the inference efficiency. 

On the other hand, the data-driven nature of the GPR imposes requirements that the dataset must be large enough, especially for high-dimensional problems. To alleviate the burden of collecting large amounts of data, incorporating some specific domain knowledge could be helpful. The physics-informed neural network (PINN) achieves great success in combining an essentially data-driven tool (i.e. deep learning) with physics knowledge \cite{UQDL088}. As for the GPR, by the insight that applying some differential operator to a Gaussian process results in another Gaussian process, \cite{UQDL075} derives the relationship between the covariance kernels of the solution and the right-hand side term for a linear PDE, then apply this result to a auto-regressive model. A subsequent study extends this idea to evolution PDEs, and predicts the solution step-by-step with the assistance of time discretisation schemes like backward Euler \cite{UQDL076}. In \cite{oUQDL012}, the authors derive the kernel for the solution of PDE given the kernel of the latent functions, as long as the Green's function for this PDE is available. Different from all the above works, in physics-informed kriging (PhIK) \cite{UQDL050}, the authors do not assume some particular form for the covariance kernel, but obtain the means and covariance from sampling a few realisations from the system of interest. Combining PhIK and multi-fidelity approach, the authors propose to construct the low-fidelity Gaussian process with PhIK instead of conventional kriging \cite{UQDL060}. In \cite{UQDL077}, the authors assume that the residual of the model follows Gaussian distribution, and derive a loss function with a physics-informed regularisation term. Among the works in this line, a vast majority will assume a fixed form of covariance form, with some hyper-parameters to be trained. This makes feasible to derive the closed form of the solution data covariance function based on some known facts. However, this will sacrifice the flexibility of the covariance kernel, because we may otherwise use powerful deep neural networks to approximate (c.f. deep kernel learning in \cite{UQDL046}). The PhIK \cite{UQDL050} and CoPhIK \cite{UQDL060} does not stick to a certain form of covariance kernel, but use MC to evaluate it instead, thus will compromise the efficiency. The work which is most similar to our proposed method could be \cite{UQDL077}, in which they involve the physical regularisation while do not exclude the possibility to use highly expressive deep kernel. We will further discuss the differences from ours and point out our advantages in section \ref{PI_GPR}, right after we reach our regularised form of loss function in equation \ref{loss}.

Recently, as deep learning becomes more and more popular, many scholars start to investigate the application of deep learning tools on the UQ field. Built upon pioneering works like \cite{UQDL004} and \cite{EEDL08}, Karumuri et al. propose variational formulations for stochastic PDEs, and use it as the loss function to train the deep neural networks \cite{EEDL03}. Besides conventional deep neural network, considering the nature of UQ problems, Bayesian deep neural networks, which is able to output a distribution instead of a simple point estimation, are also exploited in this field \cite{UQDL009}. As generative adversarial nets proposed by \cite{k002} have been proven powerful in many practical applications \cite{Gfm001, Gfm009, ke6004}, the authors propose to train a generative model in an adversarial manner, in order to match the true data distribution and model output distribution \cite{oUQDL002}.

Although GPR seems to be an alternative and competing methods to deep neural networks, the efforts to combine them together never stop. Mimicking the hierarchical structure of deep neural networks, Damianou and Lawrence develop deep Gaussian process model, in which multiple standard Gaussian process model are stacked and optimised together \cite{UQDL030}. In \cite{UQDL046}, the inputs are first mapped to the feature space by a nonlinear deep neural network, then a conventional covariance function is applied to the output of the neural network instead of inputs themselves. Denoising autoencoders \cite{UQDL068} and even ensemble of multiple deep neural networks are applied for better feature extraction in \cite{UQDL041}.

In this work, we propose a hybrid data driven-physics constrained Gaussian process regression framework. We use the GPR with deep kernel as the basis, then encode the physics constraints with Boltzmann-Gibbs distribution and derive our model by maximum likelihood (ML) approach. The model learns the deep covariance function from both the observed data and the underlying physics laws. The rest of this paper is organised as follows: in section \ref{sec:method}, we briefly review the conventional GPR model and deep kernel learning, then we derive the general form of the proposed framework. We concretise our abstract framework in a second-order PDE and conduct the numerical experiments to demonstrate the effectiveness of the proposed framework in section \ref{sec:NE}. The conclusions are presented in section \ref{sec:conclusions}.

\section{Methodology}
\label{sec:method}

\subsection{Problem settings}
Consider the following general form of a stochastic PDE
\begin{equation}\label{SPDE}
\mathcal{N}(u(\boldsymbol x); I(\omega))=g(\boldsymbol x), \boldsymbol x\in D\subset\mathbb{R}^d
\end{equation}
where $\omega\in\Omega$ is the outcome drawn from the outcome space $\Omega$, $I$ is the random variable serving as parameters of this equation (might be in some high-dimensional, or even infinite-dimensional space), $g$ is known and irrelevant to $u$, usually representing effects external to the system, and $u$ is the unknown function to be solved from this PDE. We are interested in the distribution of the solution when $\omega$ follows some probability distribution.

\subsection{Gaussian process regression and deep kernel learning}
\label{GPR}
We will give a brief review on the GPR model. For more detailed exploration on GPR, we refer interested readers to Williams and Rasmussen's celebrated text \cite{RW}.

Assume that we are provided with a dataset of $N$ labelled instances, $\{\boldsymbol{X}_i, \boldsymbol{y}_i\}_{i=1}^N$, in which $X=(\boldsymbol{x}_1, ..., \boldsymbol{x}_N)$ is the matrix stacked with $N$ multi-dimensional inputs. We seek to find an unknown underlying function $\boldsymbol{f}$ describing the data generation process. Assuming that the function outputs at any set of points follows a multivariate Gaussian distribution, with mean

\begin{equation}
\boldsymbol{\mu}=(\mu(\boldsymbol{x}_1), ..., \mu(\boldsymbol{x}_N))^T
\end{equation}

and covariance matrix

\begin{equation}
\boldsymbol{K}=
\begin{pmatrix}
k(\boldsymbol{x}_1, \boldsymbol{x}_1) & \cdots & k(\boldsymbol{x}_1, \boldsymbol{x}_N)\\
\vdots & & \vdots\\
k(\boldsymbol{x}_N, \boldsymbol{x}_1) & \cdots & k(\boldsymbol{x}_N, \boldsymbol{x}_N)
\end{pmatrix}
\end{equation}

where $\mu(\cdot)$ is a pre-defined mean function, and $k(\cdot, \cdot)$ is the covariance kernel function. In this work, we simply set $\mu(\cdot)\equiv 0$ and choose a simple squared exponential function as the covariance kernel

\begin{equation}\label{SE}
k_{SE}(\boldsymbol{x}, \boldsymbol{x}^\prime)=e^{-\frac {\Vert \boldsymbol{x}-\boldsymbol{x}^\prime\Vert}{l}}
\end{equation}

Ordinarily, the covariance kernel $k(\cdot, \cdot)$ is encoded with a set of hyper-parameters $\boldsymbol{\theta}$. Those hyper-parameters are determined through maximise the log-likelihood $\log P(\boldsymbol{y}|\boldsymbol{\theta}, \boldsymbol{X})$

\begin{equation} \label{ML}
\log P(\boldsymbol{y}|\boldsymbol{\theta}, \boldsymbol{X})\propto -(\boldsymbol{y}^T (\boldsymbol K+\sigma^2\boldsymbol I)^{-1}\boldsymbol y+\log\det{(\boldsymbol K+\sigma^2\boldsymbol I)})
\end{equation}

In this work, we adopt the deep kernel learning approach \cite{UQDL046}. In deep kernel learning, the hyper-parameters $\boldsymbol{\theta}$ appear as the parameters of the deep neural network $\boldsymbol f(\cdot;\boldsymbol{\theta})$. The covariance kernel function can be written as

\begin{equation}\label{DK}
k_{DNN}(\boldsymbol{x}, \boldsymbol{x}^\prime)=k_{SE}(\boldsymbol f(\boldsymbol{x}; \boldsymbol{\theta}), \boldsymbol f(\boldsymbol{x}^\prime;\boldsymbol{\theta}))
\end{equation}

By training the deep neural network $\boldsymbol f(\cdot;\boldsymbol{\theta})$ with regards to parameters $\boldsymbol{\theta}$, the model learns to match the observed data $\{\boldsymbol{X}_i, \boldsymbol{y}_i\}_{i=1}^N$\footnote[1]{Besides the parameters of the neural network itself, one may still add hyper-parameters to the formula of the covariance kernel function as in \cite{UQDL046}. In this work, we simply use a fixed formula ($l=2$), and train the deep neural network only to fit the data.}.

To do inference, at the unseen locations $\boldsymbol{X}^*$, the predicted distribution is a Gaussian distribution with mean

\begin{equation}\label{inference}
\boldsymbol{\mu}^*=\boldsymbol{K}(\boldsymbol{X}^*, \boldsymbol{X})(\boldsymbol{K}(\boldsymbol{X},\boldsymbol{X})+\sigma \boldsymbol{I})^{-1}\boldsymbol{y}
\end{equation}

and covariance 

\begin{equation}
\boldsymbol{K}^*=\boldsymbol{K}(\boldsymbol{X}^*, \boldsymbol{X}^*)-\boldsymbol{K}(\boldsymbol{X}^*, \boldsymbol{X})(\boldsymbol{K}(\boldsymbol{X}, \boldsymbol{X})+\sigma I)^{-1}\boldsymbol{K}(\boldsymbol{X}, \boldsymbol{X}^*)
\end{equation}

in which the meaning of expression $\boldsymbol{K}(\boldsymbol{X}, \boldsymbol{X}^{\prime})$ is the covariance matrix computed between locations $\boldsymbol{X}$ and $\boldsymbol{X}^{\prime}$, the $(i, j)$-th element of $\boldsymbol{K}(\boldsymbol{X}, \boldsymbol{X}^{\prime})$ is the covariance between the $i$-th column vector of $\boldsymbol{X}$ and the $j$-th column vector of $\boldsymbol{X}^{\prime}$.

\subsection{Proposed framework}
\label{proposed}

\subsubsection{Physics constrained deep kernel learning}
\label{PI_GPR}

The training objective function of the GPR (equation \ref{ML}) only involves the observed data, but not the physics knowledge. We propose to encode the physics constraints (e.g. PDEs) with a Boltzmann-Gibbs distribution

\begin{equation}\label{Ph}
P(\boldsymbol y|\boldsymbol X, \mathcal{P})=\frac{e^{-\beta L(\boldsymbol y, \boldsymbol X)}}{Z(\beta, \boldsymbol X, \mathcal{P})}
\end{equation}

where $\mathcal{P}$ refers to some physics constraints (e.g. PDEs), it is a random elements residing in a metric space $\mathbb{P}$. For more discussion on random elements in metric spaces, please refer to the review paper \cite{oUQDL030}. $L$ is the loss function measuring the departure of the data $(\boldsymbol y^*, \boldsymbol X^*)$ from this physics constraints. In the case that the physics constraints is represented as PDEs, $L$ could be square of residuals \cite{FPDL06}, or in a variational form \cite{EEDL08}. $\beta$ is a pre-defined constant, representing the strength of the physics constraints. We can assign large value to $\beta$ if we are more confident with the physics constraints rather than the observed data. $Z$ is a normalisation constant in order to make $P(\boldsymbol y^*|\boldsymbol X^*, \mathcal{P})$ a qualified probability distribution. We will see later that the value of this constant is of no practical importance.

\begin{figure}[H]
  \centering
  \includegraphics[scale=0.15]{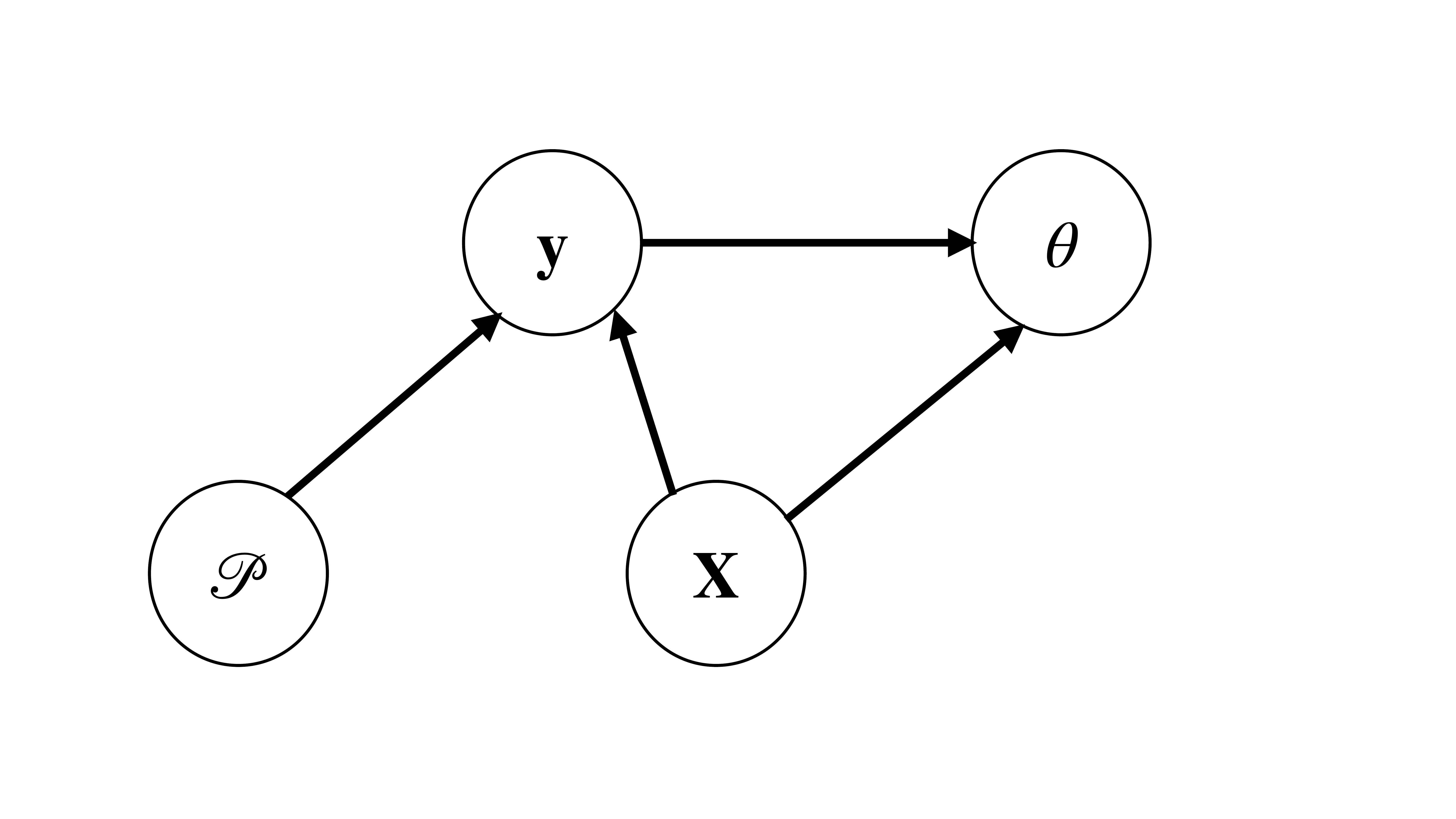}
  \caption{Bayesian network representing the relationships among $\mathcal{P}$, $\boldsymbol X$, $\boldsymbol y$ and $\boldsymbol{\theta}$}
  \label{fig:BN}
\end{figure}

In our proposed framework, there are 4 random elements to be considered, i.e. $\mathcal{P}$, $\boldsymbol X$, $\boldsymbol y$ and $\boldsymbol{\theta}$. Some of them retain cause-effect relationships, while some are independent from each other. For example, the inputs $\boldsymbol X$ and the physical law $\mathcal{P}$ together decide the value of the output $\boldsymbol y$, but which inputs $\boldsymbol X$ are sampled are independent from the underlying physical law $\mathcal{P}$, because the sampling method are selected by us humans. We use a Bayesian network to depict such relationships, see figure \ref{fig:BN}.

Now we will derive the training objective of our model. First, it is desired that the evidence $P(\boldsymbol{y}|\boldsymbol{\theta}, \boldsymbol{X}, \mathcal{P}=\mathcal{P}^*)$, in which $\mathcal{P}^*$ is the physical law governing the system (e.g. the PDE through which the uncertainty propagates), is large. This evidence is conditioned on both the inputs $\boldsymbol X$ and the physical law $\mathcal{P}^*$, meaning that the underlying physics knowledge is taken into account. Also, it is conditioned on the model parameters $\boldsymbol{\theta}$, meaning that we assume a Gaussian structure on data, in which the outputs are correlated according to the proximity of their inputs. This evidence term represents the "physics-constrained" aspect of the model. Secondly, since we are provided with some paired data, regardless the model structure or the physical law, we also want the model to fit the known data, thus the term $P(\boldsymbol{y}|\boldsymbol{X})$ should also be included. This term represents the "data-driven" aspects of our model. As a result, we will minimise the negative log-likelihood of the product of those two terms

\begin{equation}
\mathcal{L}=-\log [P(\boldsymbol{y}|\boldsymbol{\theta}, \boldsymbol{X}, \mathcal{P}=\mathcal{P}^*)P(\boldsymbol{y}|\boldsymbol{X})] \label{ori_L}
\end{equation}

The figure \ref{fig:BN} facilitate us to decompose expression \ref{ori_L} into tractable terms. Based on \ref{fig:BN}, we write the joint distribution to be

\begin{equation}
P(\boldsymbol{y}, \boldsymbol{\theta}, \boldsymbol{X}, \mathcal{P}=\mathcal{P}^*)=P(\mathcal{P}=\mathcal{P}^*)P(\boldsymbol{X})P(\boldsymbol{y}|\mathcal{P}=\mathcal{P}^*, \boldsymbol{X})P(\boldsymbol{\theta}|\boldsymbol{X}, \boldsymbol{y})
\end{equation}

Then the original product in \ref{ori_L} can be written as

\begin{equation}
\begin{aligned}
P(\boldsymbol{y}|\boldsymbol{\theta}, \boldsymbol{X}, \mathcal{P}=\mathcal{P}^*)P(\boldsymbol{y}|\boldsymbol{X})
&=\frac{P(\boldsymbol{y}, \boldsymbol{\theta}, \boldsymbol{X}, \mathcal{P}=\mathcal{P}^*)}{P(\boldsymbol{\theta}, \boldsymbol{X}, \mathcal{P})}P(\boldsymbol{y}|\boldsymbol{X})\\
&=\frac{P(\mathcal{P}=\mathcal{P}^*)P(\boldsymbol{X})P(\boldsymbol{y}|\mathcal{P}=\mathcal{P}^*, \boldsymbol{X})P(\boldsymbol{\theta}|\boldsymbol{X}, \boldsymbol{y})}{P(\boldsymbol{\theta}, \boldsymbol{X}, \mathcal{P}=\mathcal{P}^*)}P(\boldsymbol{y}|\boldsymbol{X})\\
&=\frac{P(\mathcal{P}=\mathcal{P}^*)P(\boldsymbol{X})P(\boldsymbol{y}|\mathcal{P}=\mathcal{P}^*, \boldsymbol{X})P(\boldsymbol{X}, \boldsymbol{y}|\boldsymbol{\theta})P(\boldsymbol{\theta})}{P(\boldsymbol{\theta}, \boldsymbol{X}, \mathcal{P}=\mathcal{P}^*)P(\boldsymbol{X}, \boldsymbol{y})}P(\boldsymbol{y}|\boldsymbol{X})\\
&=\frac{P(\mathcal{P}=\mathcal{P}^*)P(\boldsymbol{X})P(\boldsymbol{y}|\mathcal{P}=\mathcal{P}^*, \boldsymbol{X})P(\boldsymbol{y}|\boldsymbol{X}, \boldsymbol{\theta})P(\boldsymbol{X}|\boldsymbol{\theta})P(\boldsymbol{\theta})}{P(\boldsymbol{\theta}, \boldsymbol{X}, \mathcal{P})P(\boldsymbol{X}, \boldsymbol{y})}P(\boldsymbol{y}|\boldsymbol{X})\\
&=P(\boldsymbol{y}|\mathcal{P}=\mathcal{P}^*, \boldsymbol{X})P(\boldsymbol{y}|\boldsymbol{X}, \boldsymbol{\theta})\frac{P(\boldsymbol{X})P(\boldsymbol{y}|\boldsymbol{X})}{P(\boldsymbol{X}, \boldsymbol{y})}\frac{P(\mathcal{P}=\mathcal{P}^*)P(\boldsymbol{X}|\boldsymbol{\theta})P(\boldsymbol{\theta})}{P(\boldsymbol{\theta}, \boldsymbol{X}, \mathcal{P}=\mathcal{P}^*)}\text{ (rearrange the terms)}\\
&=P(\boldsymbol{y}|\mathcal{P}=\mathcal{P}^*, \boldsymbol{X})P(\boldsymbol{y}|\boldsymbol{X}, \boldsymbol{\theta})\frac{P(\boldsymbol{\theta}, \boldsymbol{X})}{P(\boldsymbol{\theta}, \boldsymbol{X}|\mathcal{P}=\mathcal{P}^*)}\label{main_derivation}
\end{aligned}
\end{equation}

In the third equality we use Bayesian rule $P(\boldsymbol{\theta}|\boldsymbol{X}, \boldsymbol{y})=\frac{P(\boldsymbol{X}, \boldsymbol{y}|\boldsymbol{\theta})P(\boldsymbol{\theta})}{P(\boldsymbol{X}, \boldsymbol{y})}$. Here we need to assign a prior to $\mathcal{P}$. We will impose an infinitely strong prior on $\mathcal{P}$

\begin{equation}
P(\mathcal{P}=z)=\left\{
\begin{array}{rcl}
1 & & {z=\mathcal{P}^*}\\
0 & & {\text{otherwise}}
\end{array}\right.
\end{equation}

Use the law of total probability, we get

\begin{equation}
\begin{aligned}
P(\boldsymbol{\theta}, \boldsymbol{X})
&=\sum_{z\in\mathbb{P}}P(\boldsymbol{\theta}, \boldsymbol{X}|\mathcal{P}=z)P(\mathcal{P}=z)\\
&=P(\boldsymbol{\theta}, \boldsymbol{X}|\mathcal{P}=\mathcal{P}^*)
\end{aligned}
\end{equation}

Plug into the last line of equation \ref{main_derivation}, the last fractional number cancelled out, we reach the objective function

\begin{equation}
\begin{aligned}
\mathcal{L}&=-\log [P(\boldsymbol{y}|\boldsymbol{\theta}, \boldsymbol{X}, \mathcal{P}=\mathcal{P}^*)P(\boldsymbol{y}|\boldsymbol{X})]\\
&=-\log P(\boldsymbol{y}|\mathcal{P}=\mathcal{P}^*, \boldsymbol{X})-\log P(\boldsymbol{y}|\boldsymbol{X}, \boldsymbol{\theta})
\end{aligned}
\end{equation}

Using equation \ref{Ph}, we have $P(\boldsymbol{y}|\mathcal{P}=\mathcal{P}^*, \boldsymbol{X})P(\boldsymbol{y}|\boldsymbol{X}, \boldsymbol{\theta})=\beta L^*(\boldsymbol y, \boldsymbol X)+Z(\beta, \boldsymbol X, \mathcal{P}^*)$, where $L^*$ is the loss corresponding to physical law $\mathcal{P}^*$. Notice that $Z(\beta, \boldsymbol X, \mathcal{P}^*)$ is a constant w.r.t. the model parameter $\boldsymbol{\theta}$, thus it can be ignored in the objective function. For the second term, in GPR models, $-\log P(\boldsymbol{y}|\boldsymbol{X}, \boldsymbol{\theta})=\boldsymbol{y}^T (\boldsymbol K+\sigma^2\boldsymbol I)^{-1}\boldsymbol y+\log\det{(\boldsymbol K+\sigma^2\boldsymbol I)}$. We can finally write out the training objective of the hybrid model

\begin{equation}
\mathcal{L}=\beta L^*(\boldsymbol y, \boldsymbol X)+\boldsymbol{y}^T (\boldsymbol K+\sigma^2\boldsymbol I)^{-1}\boldsymbol y+\log\det{(\boldsymbol K+\sigma^2\boldsymbol I)}\label{loss}
\end{equation}

We now arrive at a pleasant form of the objective function equation \ref{loss}, which can be regarded as the sum of the model log-likelihood and the regularisation. In \cite{UQDL077} and a series of subsequent works \cite{UQDL061, UQDL078, UQDL079}, the authors also reach a similar form. They propose a GPR framework with a physics-regularisation term. However, in \cite{UQDL077}, the residual (in those works, the latent force) is supposed to be closed to zero (or at least some known, fixed value), thus the variational form of the PDE cannot be used, because its minimum is unknown, the only thing we know about the variational form is that its minimiser is the solution of that PDE. In many cases, we may prefer the variational form while defining the PDE related loss function, because in this form usually the highest order of derivatives appeared is decreased by one. Take the Poisson equation as an example

\begin{equation}
-\Delta u=g\text{ in } U
\end{equation}

If we take the variational form to design the loss function, it will be

\begin{equation}
L=\int_U\frac12\Vert\nabla u\Vert^2-ugdx
\end{equation}\cite{LW}.

We can see that the second-order derivative disappears in the variational form. This property of the variational form is important to numerical stability, since numerically calculating the derivatives usually involves computing the differences of two close numbers and the division with a very small divisor, and the situation is worse in higher-order case. In our proposed framework, we do not impose this restriction on the PDE related loss. The users can freely choose any form for this part of the loss, as long as it can truly reflect how well the data fits the physical law.

\subsubsection{Regression on the coefficient space}

Most of the covariance kernels used in the GPR assume that when the input variables are closed in space, they will be closed related. However, this is not true for the solution field of PDEs, since the weak solution of PDEs might be discontinuous even when the coefficients are smooth. To address this issue, we propose to use the (random) coefficients as the input data $\boldsymbol X$, i.e., the dataset $\boldsymbol X=\{I(\omega_i)\}_{i=1}^N$, $\omega_i\in \Omega$. As for the output, we expect the method to output the solution over the whole domain $D$. That is impossible since the output is in a finite-dimensional space, but we may use some colocation method to discretise $D$ as in \cite{UQDL004}. Assuming that we adopt a simple uniform grid, then the output of the regression will then be in $\mathbb{R}^{n_1\times n_2\times\cdots\times n_d}$, where $n_i$ is the number of grid points along dimension $i$. The equation \ref{loss} is applicable when the target is one-dimensional. We assume that the different output entries are independent following \cite{UQDL021}, then the log-likelihood will be

\begin{equation}\label{Ind}
\log P(\boldsymbol{y}|\boldsymbol{\theta}, \boldsymbol{X})=\sum_{i=1}^{n_1n_2\cdots n_d} \log P(\boldsymbol{y}_i|\boldsymbol{\theta}, \boldsymbol{X})
\end{equation}

where $\boldsymbol y_i\in\mathbb{R}^{N}$ is the $i$-th elements of the flattened version of the target, and $N$ is the number of the data points.

\subsubsection{Stochastic inducing points}
\label{subsec:SIP}

In our proposed framework, in order to calculate the physics-informed loss, we need to do GPR inference every training step. As a result, it is critical to use an efficient inference algorithm. Unfortunately, one critical shortcoming of the GPR is the time complexity required for evaluating $(\boldsymbol{K}(\boldsymbol{X},\boldsymbol{X})+\sigma \boldsymbol{I})^{-1}$ in equation \ref{inference} during inference time. A common way to do this is through Cholesky decomposition, with cubic time complexity with regard to the number of data points. There are a number of works aiming at alleviating this difficulty \cite{oUQDL007, oUQDL006, UQDL055}. Here, to fit the nature of deep learning algorithm, we propose a simple yet useful scheme to reduce the burden of inverting a huge matrix. Let $s$ be the training batch size, let $\{\boldsymbol x_i, \boldsymbol y_i\}_{i=1}^s$ be a batch of the labelled data, let $\{S_k, S_{uk}\}$ be a split of the set $S:={1,2,...,s}$, in other words, $S_k\cap S_{uk}=\emptyset$, and $S_k\cup S_{uk}=S$. Then, we divide the batch into 2 parts: $B_k:=\{\boldsymbol x_i, \boldsymbol y_i\}_{i\in S_k}$ and $B_{uk}:=\{\boldsymbol x_i, \boldsymbol y_i\}_{i\in S_{uk}}$. We will pretend that we do not know the label of the data in $B_{uk}$, and use the inference algorithm of the GPR (see equation \ref{inference}) to evaluate the label of the data in $B_{uk}$, we denote them (in conjunction with their input) with $I:=\{\boldsymbol x_i, \boldsymbol {\hat y}_i\}_{i\in S_{uk}}$. Then, we will be able to use $I$ to evaluate the physics-informed part of the loss.

\subsubsection{Overall training procedure}

Here we formulate the above thoughts mathematically. We stack the input and output vectors in $B$, $B_k$ and $B_{uk}$ to be matrix, say $\boldsymbol X, \boldsymbol Y, \boldsymbol X_k, \boldsymbol Y_k, \boldsymbol X_{uk}, \boldsymbol Y_{uk}$ (e.g. the columns of matrix $\boldsymbol X_k$ are vectors in $B_k$). Consider the $i$-th entry of the outputs $\boldsymbol Y$ and $\boldsymbol Y_k$, they are the $i$-th row of the matrix $\boldsymbol Y$ and $\boldsymbol Y_k$, denote it by $\boldsymbol y_i$ and $\boldsymbol y^{(k)}_i$, respectively. Then we can infer the $i$-th entry of the outputs for inputs $\boldsymbol X_{uk}$ through the equation \ref{inference}:

\begin{equation}\label{I}
\boldsymbol {\hat{y}}_i=K(\boldsymbol X_{uk}, \boldsymbol X_k)(K(\boldsymbol X_{k}, \boldsymbol X_k)+\sigma \boldsymbol I)^{-1}\boldsymbol y_i
\end{equation}

Then, plug into the equation \ref{loss}, we obtain the likelihood for the $i$-th output entry:

\begin{equation}\label{EleL}
\mathcal{L}_i=-(\boldsymbol{y}_i^T (\boldsymbol K(\boldsymbol X, \boldsymbol X)+\sigma^2\boldsymbol I)^{-1}\boldsymbol y_i+\log\det{(\boldsymbol K(\boldsymbol X, \boldsymbol X)+\sigma^2\boldsymbol I)})-\beta L(\boldsymbol {\hat{y}}_i, \boldsymbol X_{uk})
\end{equation}

Using equation \ref{Ind}, we obtain the final loss function:

\begin{equation}\label{FL}
\mathcal{L}=\sum_i \mathcal{L}_i
\end{equation}

One particularly desirable property of this training procedure is that it is online. Nowadays, sometimes the complete datasets might be stored in low-cost tertiary storage like hard disk drives or even tapes. In such scenarios, random access is impossible or extremely expensive. In our scheme, all data items are read one-by-one, which means that the training can work efficiently in such cases.

We summarise the training procedure of our proposed framework in algorithm \ref{alg: Train}.

\begin{algorithm}[!t]
\caption{Training Procedure}
\label{alg: Train}
\hspace*{0.02in}{\bf Input: } 
Batched dataset $\mathcal{D}$, number of epochs $E$. \\
\begin{algorithmic}[1]
\State i=0
\While{$i<E$}
	\For {Batch $B=\{\boldsymbol x_i, \boldsymbol y_i\}_{i=1}^s$ in $\mathcal{D}$}
		\State Split $B$ into $B_k$ and $B_{uk}$
		\State Formulate $\boldsymbol X, \boldsymbol Y, \boldsymbol X_k, \boldsymbol Y_k, \boldsymbol X_{uk}, \boldsymbol Y_{uk}$
		\State Infer $\boldsymbol {\hat{y}}_i$ with equation \ref{I}
		\State Calculate the loss $\mathcal{L}$ using equation \ref{EleL} and equaiton \ref{FL}
		\State Update the network parameter using some optimiser, like Adam, according to $\mathcal{L}$.
		\State i++
	\EndFor
\EndWhile
\end{algorithmic}
\end{algorithm}

\section{Numerical experiments}
\label{sec:NE}

To demonstrate the effectiveness of the proposed framework, we conduct the numerical experiments on the steady-state solution for the diffusion equation

\begin{equation}\label{DE}
\nabla\cdot(D(x)\nabla u(x))=0, x\in U
\end{equation}

where $D$ is a random field representing the diffusivity of the medium, and $U=[0,1]\times[0,1]$ is the spatial domain. We assume Dirichlet and Neumann boundary conditions:

\begin{equation}
\begin{aligned}
u(x)=1, x\in\{0\}\times[0,1]\\
u(x)=0, x\in\{1\}\times[0,1]\\
D\nabla u(x)\cdot\boldsymbol n=0, x\in\partial U
\end{aligned}
\end{equation}

The physics constraint loss $L$ in equation \ref{Ph} is defined in a variational form in this experiment:

\begin{equation}\label{Ph_loss_NE}
L(u)=\int_U (\frac12 D\nabla u\cdot\nabla u)dx+\int_{\{0\}\times[0,1]} (u-1)^2 ds+\int_{\{1\}\times[0,1]} u^2 ds
\end{equation}

where $\int_{\{0\}\times[0,1]} (u-1)^2 ds$ and $\int_{\{1\}\times[0,1]} u^2 ds$ are line integrals enforcing the Dirichlet boundary condition.

In this work, we use the dataset provided in \cite{UQDL002}. The Gaussian random field (GRF) is generated through Karhunen-Lo\'{e}ve expansion, sampled as 32$\times$32 or 64$\times$64 images in this dataset, and the random field and the solution field $u$ is paired \cite{UQDL002}. 

Since in this dataset the $u$ is discretised, which means it is represented with the function values at grids points, we adapt the equation \ref{Ph_loss_NE} to be:

\begin{equation}
L(\boldsymbol {\hat{y}}, \boldsymbol x)=\frac{1}{N_{i}N_j}\sum_{i,j} (\frac12 \boldsymbol x(i,j)\nabla \boldsymbol {\hat{y}}(i,j)\cdot\nabla \boldsymbol {\hat{y}}(i,j))+\frac{1}{N_{i}}\sum_{i} (\boldsymbol {\hat{y}}(i,0)-1)^2 ds+\frac{1}{N_{i}}\sum_{i} (\boldsymbol {\hat{y}}(i,N_i-1))^2 ds
\end{equation}

given a single pair of inferred solution $\boldsymbol {\hat{y}}$ and input random field $\boldsymbol{x}$. $N_i$ and $N_j$ are number of rows and columns in the grids (for simplicity, we assume uniform rectangular grid), and $\boldsymbol x(i,j)$, $\boldsymbol {\hat{y}}(i,j)$ and $\nabla \boldsymbol {\hat{y}}(i,j)$ are function values or gradients on the grid point $(i,j)$. One may use some numerical differentiation scheme to calculate $\nabla \boldsymbol {\hat{y}}$, in this work, following the appendix A of \cite{UQDL002}, we apply the Sobel filter with some corrections near boundary.

\subsection{Exponential of Gaussian random field}
\label{GRF}

\begin{figure}[htbp]
  \centering
  \includegraphics[scale=0.5]{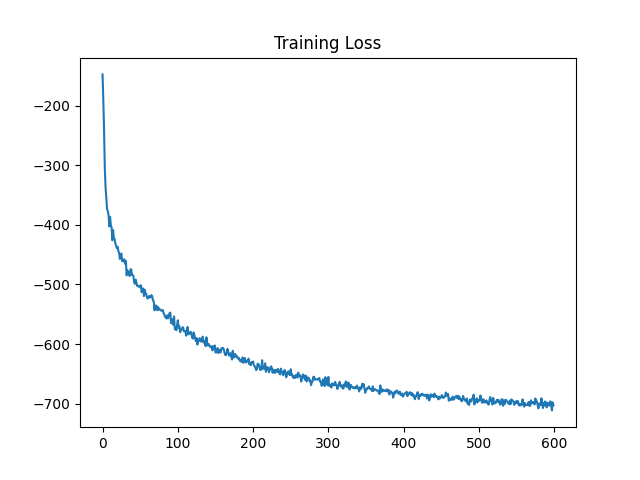}
  \caption{Training loss vs. epoch in GRF experiments.}
  \label{fig:Training_loss}
\end{figure}

\begin{figure}[htbp]
  \centering
  \includegraphics[scale=0.5]{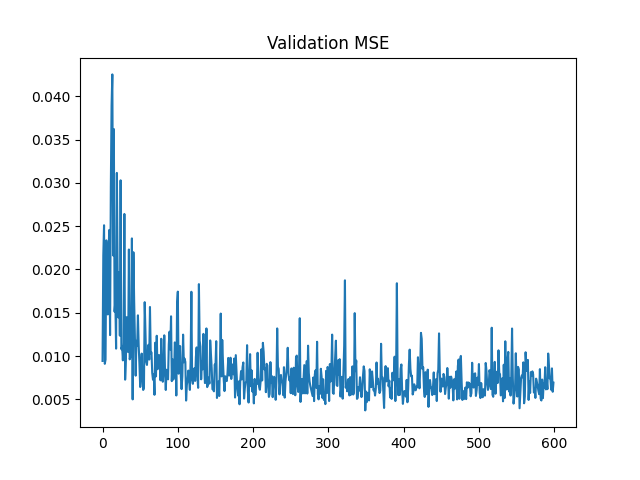}
  \caption{validation mean squared error vs. epoch in GRF experiments.}
  \label{fig:Val}
\end{figure}

We first test out our framework in the exponential of the GRF. In the exponential of the GRF, the exponents of values follow a Gaussian process (GP), i.e.

\begin{equation}
D=e^r, r\sim GP(0, k_{SE}(\cdot, \cdot))
\end{equation}

where $k_{SE}(\cdot, \cdot)$ is the covariance kernel (see equation \ref{SE}). In this dataset, $l=0.2$. The random fields and the solutions are discretised into 32$\times$32 grid.

We implement the framework with TensorFlow \cite{TF}. Follows \cite{UQDL041}, we apply a denoising autoencoder \cite{UQDL068} as the neural network used in deep kernel learning \cite{UQDL046}. Denote the encoder with $\boldsymbol f_E$ and the decoder as $\boldsymbol f_D$, then the deep kernel (see equation \ref{DK}) will be

\begin{equation}
k_{DNN}(\boldsymbol{x}, \boldsymbol{x}^\prime)=k_{SE}(\boldsymbol f_E(\boldsymbol{x}; \boldsymbol{\theta}), \boldsymbol f_E(\boldsymbol{x}^\prime;\boldsymbol{\theta}))
\end{equation}

For autoencoders, we need to add an additional squared $L^2$ loss \cite{DL} to the original loss function (equation \ref{loss})

\begin{equation}\label{DAE_loss}
\mathcal{L}_{DAE}=\mathcal{L}+\gamma\Vert \boldsymbol x-\boldsymbol f_D(\boldsymbol f_E(\boldsymbol x))\Vert^2
\end{equation}

where $\gamma$ is a user-defined hyper-parameter.

We apply the Adam optimiser \cite{UQDL089} to train the autoencoder. We train our model on 900 realisations of $D$ for 600 epochs, while using another 100 realisations to validate the model. We calculate the mean square error (MSE) between the model output and the true solution on the validation set after each epoch, and the model with the lowest MSE will be kept for testing. For testing, another 10000 realisations of $D$, which is neither in training set nor in validation set, are used to test the model accuracy. Note that although 10000 realisations seem to be a large number, those data is not used to update the network parameter. Only 900+100=1000 realisations are used for the search of the best network parameters. In other words, this model works on a relatively small dataset. More technical details of the network structure and training process used in the experiments in this paper are present in the appendix \ref{sec:App}.

Figure \ref{fig:Training_loss} shows the trends of the training loss value (equation \ref{DAE_loss}). We can see that the training loss values drops quickly at the first 200 epochs and keeps relatively stable after 500 epochs. This implies the good convergence of the model.

Figure \ref{fig:Val} shows the MSE on the validation set. Despite the fluctuation at the very beginning, the MSE on the unseen validation soon decreases, and keep being near 0.05. 

\begin{figure}[htbp]
  \centering
  \subfigure{
  \includegraphics[scale=0.7]{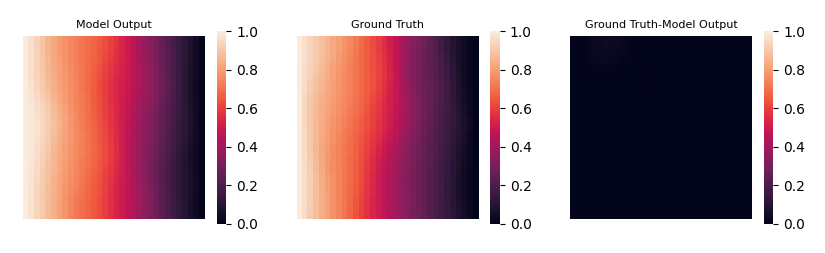}
  }
  \subfigure{
  \includegraphics[scale=0.7]{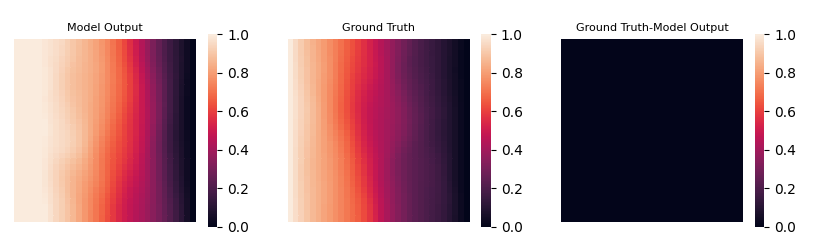}
  }
  \subfigure{
  \includegraphics[scale=0.7]{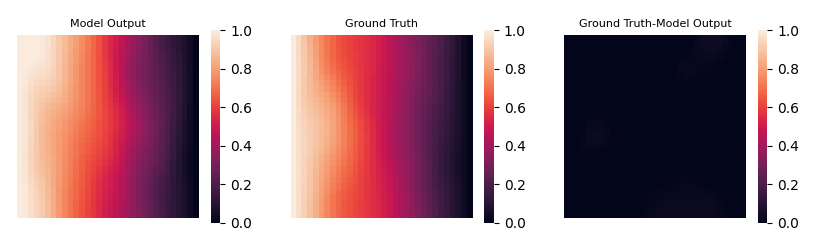}
  }
  \caption{Some instances on the test set in GRF experiment.}
  \label{fig:Test}
\end{figure}

After the training has finished and the checkpoint with the best validation MSE is recorded, we test our model on a test set, which is disjoint to both training and validation set. Figure \ref{fig:Test} shows a few instances on the test set. Each row corresponds to a realisation of $D$, and consists of the model output, the ground truth and the differences between them. Note that these data instances are never seen by the model before, neither during training nor validation. This shows that the model does not only memorise the training examples, but also learns some meaningful structure underlying the data. Thus it is able to correctly generalise to new data and make accurate predictions.

\begin{figure}[htbp]
  \centering
  \subfigure{
  \includegraphics[scale=0.4]{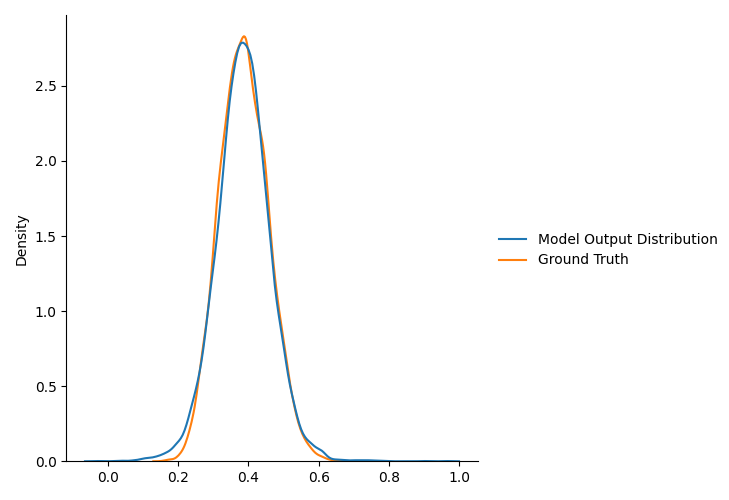}
  }
  \subfigure{
  \includegraphics[scale=0.4]{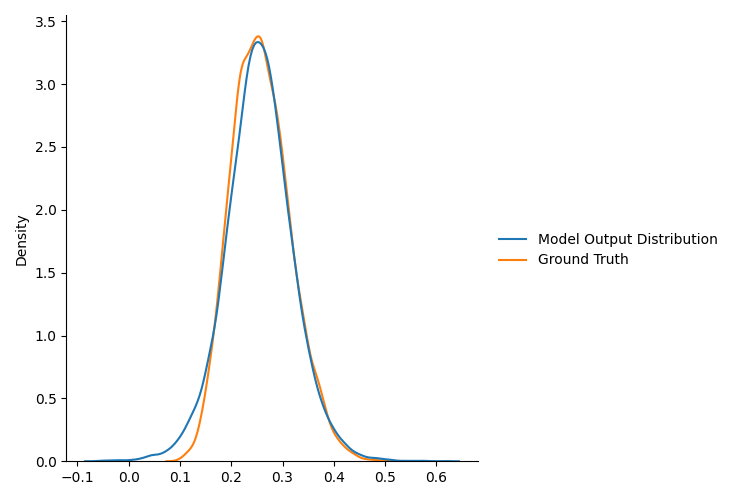}
  }
  \subfigure{
  \includegraphics[scale=0.4]{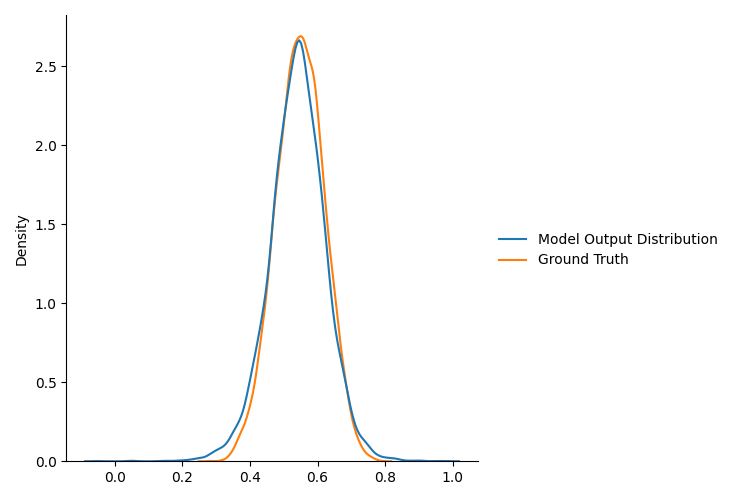}
  }
  \subfigure{
  \includegraphics[scale=0.4]{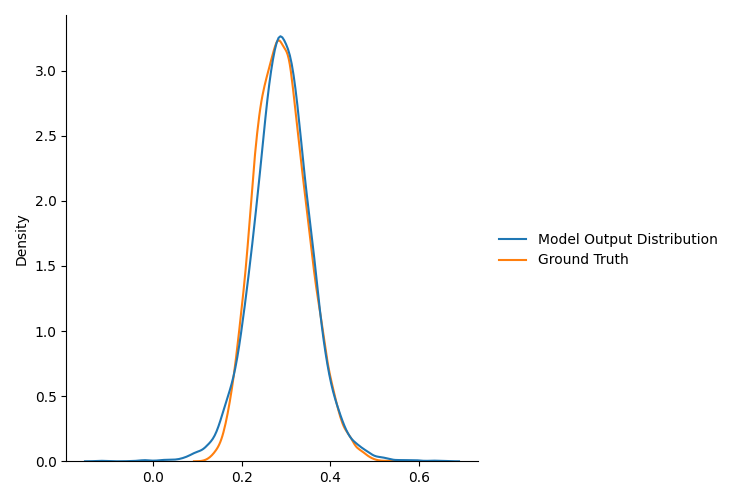}
  }
  \caption{Distribution on points (0.59375, 0.21875) (top left-hand corner), (0.71875, 0.21875) (top right-hand corner), (0.71875, 0.4375) (bottom left-hand corner) and (0.6875, 0.3125) (bottom right-hand corner).}
  \label{fig:Dist}
\end{figure}

Above results show the model's capability as a deterministic surrogate. When a realisation of the random filed is sampled, the model is able to accurately estimate the solution corresponding to this realisation. In Figure \ref{fig:Dist}, we assemble the model's output on the whole test set, and compare with the ground truth distribution of the output at a few spatial points. This shows that the model can correctly propagate the uncertainty and predict the probability distribution of the solution.

Those results show that the model is mature. Comparing with the dimensionality of the data (both input and output are of size 32$\times$32=1024), we only need a small amount of data (1000 data instances) to train it, despite the fact that conventional deep learning performs poorly on small data set because it overfits the training set but fails to generalise to new data.

\subsection{Channelised field}
\label{CF}

\begin{figure}[htbp]
  \centering
  \includegraphics[scale=0.5]{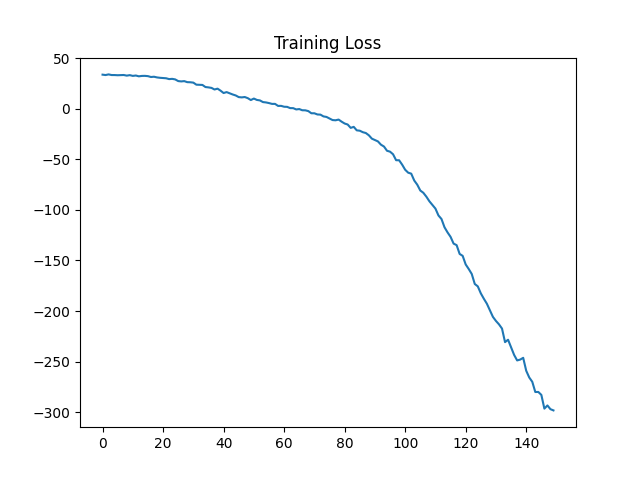}
  \caption{Training Loss vs. Epoch in Channelised Field Experiment.}
  \label{fig:Training_loss_64}
\end{figure}

\begin{figure}[htbp]
  \centering
  \includegraphics[scale=0.5]{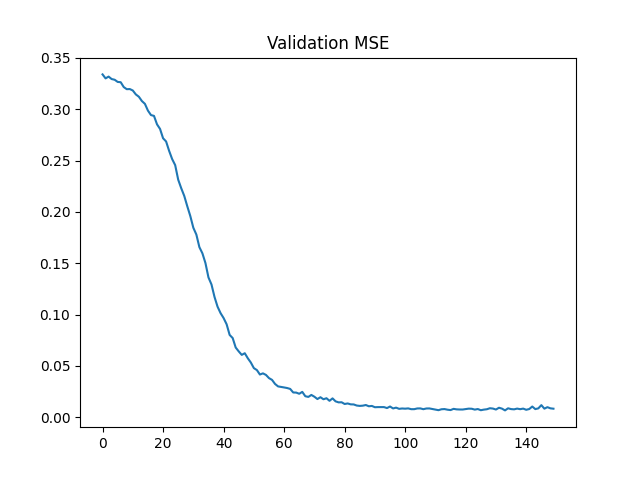}
  \caption{Validation mean squared error vs. epoch in channelised field experiment.}
  \label{fig:Val_64}
\end{figure}

\begin{figure}[htbp]
  \centering
  \subfigure{
  \includegraphics[scale=0.7]{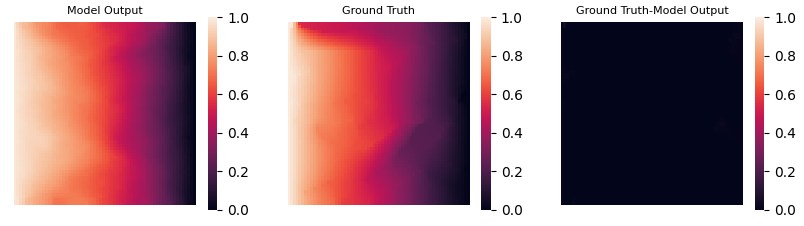}
  }
  \subfigure{
  \includegraphics[scale=0.7]{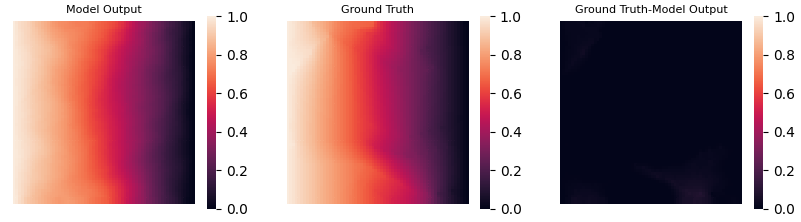}
  }
  \subfigure{
  \includegraphics[scale=0.7]{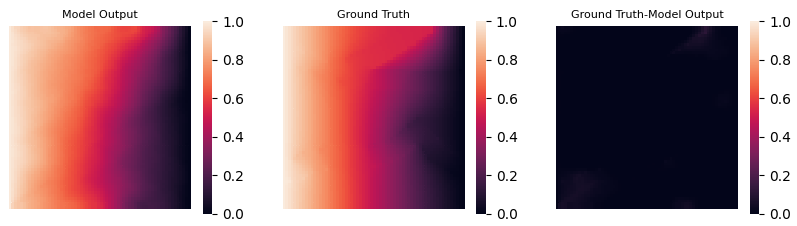}
  }
  \caption{Some instances on the test set in channelised field experiment.}
  \label{fig:Test_64}
\end{figure}

\begin{figure}[htbp]
  \centering
  \subfigure{
  \includegraphics[scale=0.4]{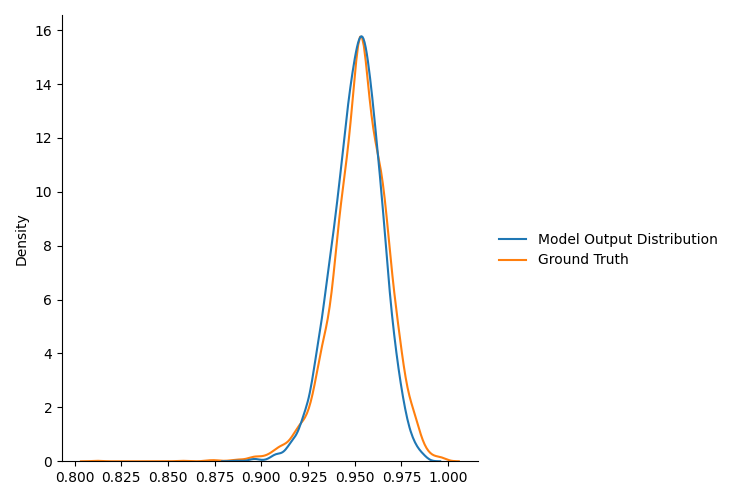}
  }
  \subfigure{
  \includegraphics[scale=0.4]{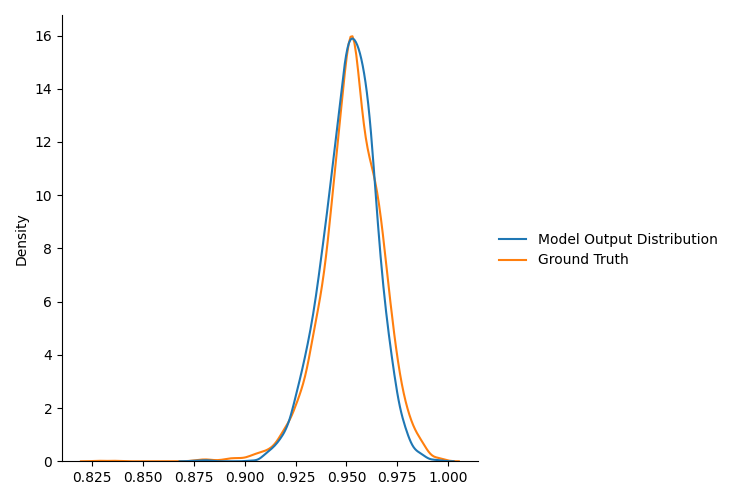}
  }
  \subfigure{
  \includegraphics[scale=0.4]{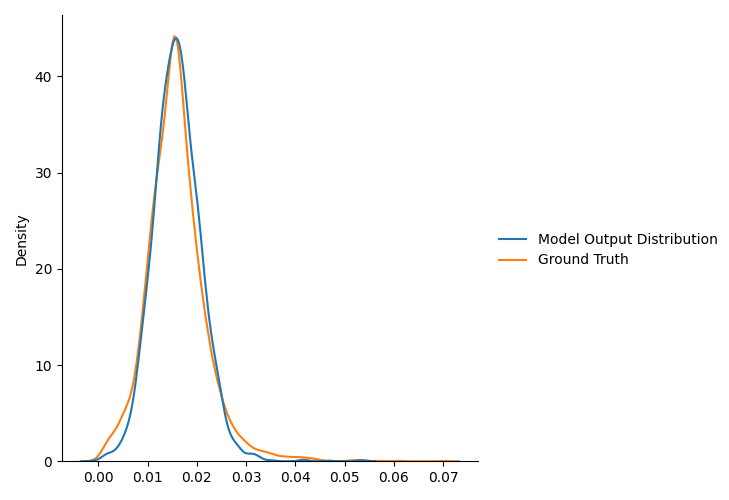}
  }
  \subfigure{
  \includegraphics[scale=0.4]{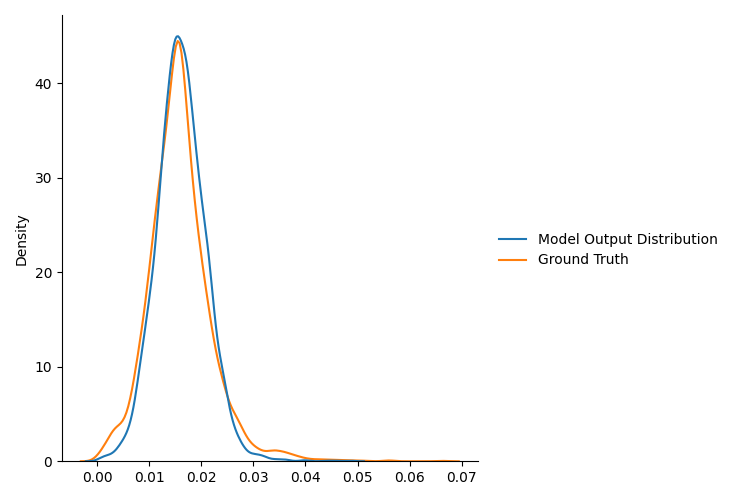}
  }
  \caption{Distribution on points (0.4375, 0.03125) (top left-hand corner), (0.46875, 0.03125) (top right-hand corner), (0.484375, 0.953125) (bottom left-hand corner) and (0.5, 0.953125) (bottom right-hand corner).}
  \label{fig:Dist_64}
\end{figure}

We also test out our method on channelised permeability field. This dataset contains cropped patches of size 64$\times$64 \cite{UQDL002}. The governing equation is still equation \ref{DE}. Similar to the case in section \ref{GRF}, we use a denoising autoencoder architecture, with adaptive modifications for new structure of the data. The dimensionality of this dataset is larger comparing with the GRF dataset in section \ref{GRF}. However, the data available for training, validation and testing is less. In this experiment, we use only 450 data instances for training, another 62 data instances for validation. 4096 data instances, which is not seen by the model during training or validation are used for testing. The model is trained for 150 epochs, and the best snapshot (with regards to the validation set) is stored for testing.

As in the GRF experiment present in section \ref{GRF}, the trends of the training loss value and the MSE on the validation set are shown in figure \ref{fig:Training_loss_64} and figure \ref{fig:Val_64}, respectively. The flat region in figure \ref{fig:Val_64} indicates the convergence of the deep model. Although it seems that the training loss is still decreasing near the end of the training, it is indeed overfitting, which is prone to happen especially in small dataset scenario. That means our model matures quickly.

The same as in section \ref{GRF}, we record the model parameters with the lowest validation MSE, and test it on a testing set of 4096 data instances unseen during training and validation. Figure \ref{fig:Test_64} shows the comparison between the model output and the ground truth for some instances in the testing set.

Figure \ref{fig:Dist_64} shows the comparison between the model output and the ground truth distribution over the whole testing set at certain points in the spatial domain.

\section{Conclusions}
\label{sec:conclusions}

In this study, we propose a hybrid data driven-physics constrained Gaussian process regression framework. We incorporate physics knowledge into the Bayesian framework of the GPR, which is essentially a data-driven machine method. The proposed model is able to learn the deep covariance function from both the observed data and the underlying physics laws. By doing this, the data amount needed to train this model to convergence becomes very small, comparing with the number of dimensions of the data in the problem.

For future works, one might try out this framework as a surrogate on other UQ applications, not limited to stochastic PDEs. Also, more concrete theoretical guarantees need to be studied on some aspects of the proposed framework, like the stochastic inducing points. More powerful and sophistical deep neural network architecture can be utilised to further improve the efficiency and accuracy.

\bibliographystyle{siam}
\bibliography{references.bib}

\begin{thebibliography}{10}

\bibitem{TF}
{\sc M.~Abadi, A.~Agarwal, P.~Barham, E.~Brevdo, Z.~Chen, C.~Citro, G.~S.
  Corrado, A.~Davis, J.~Dean, M.~Devin, S.~Ghemawat, I.~Goodfellow, A.~Harp,
  G.~Irving, M.~Isard, Y.~Jia, R.~Jozefowicz, L.~Kaiser, M.~Kudlur,
  J.~Levenberg, D.~Man\'{e}, R.~Monga, S.~Moore, D.~Murray, C.~Olah,
  M.~Schuster, J.~Shlens, B.~Steiner, I.~Sutskever, K.~Talwar, P.~Tucker,
  V.~Vanhoucke, V.~Vasudevan, F.~Vi\'{e}gas, O.~Vinyals, P.~Warden,
  M.~Wattenberg, M.~Wicke, Y.~Yu, and X.~Zheng}, {\em {TensorFlow}: Large-scale
  machine learning on heterogeneous systems}, 2015.
\newblock Software available from tensorflow.org.

\bibitem{UQDL021}
{\sc I.~Bilionis and N.~Zabaras}, {\em Multi-output local gaussian process
  regression: Applications to uncertainty quantification}, Journal of
  Computational Physics, 231 (2012), pp.~5718--5746.

\bibitem{UQDL014}
{\sc I.~Bilionis, N.~Zabaras, B.~A. Konomi, and G.~Lin}, {\em Multi-output
  separable gaussian process: Towards an efficient, fully bayesian paradigm for
  uncertainty quantification}, Journal of Computational Physics, 241 (2013),
  pp.~212--239.

\bibitem{Gfm009}
{\sc L.~Chen, S.~Dai, C.~Tao, H.~Zhang, Z.~Gan, D.~Shen, Y.~Zhang, G.~Wang,
  R.~Zhang, and L.~Carin}, {\em Adversarial text generation via
  feature-mover\textquotesingle s distance}, in Advances in Neural Information
  Processing Systems 31, S.~Bengio, H.~Wallach, H.~Larochelle, K.~Grauman,
  N.~Cesa-Bianchi, and R.~Garnett, eds., Curran Associates, Inc., 2018,
  pp.~4666--4677.

\bibitem{UQDL030}
{\sc A.~Damianou and N.~D. Lawrence}, {\em Deep {G}aussian processes}, in
  Proceedings of the Sixteenth International Conference on Artificial
  Intelligence and Statistics, C.~M. Carvalho and P.~Ravikumar, eds., vol.~31
  of Proceedings of Machine Learning Research, Scottsdale, Arizona, USA, 29
  Apr--01 May 2013, PMLR, pp.~207--215.

\bibitem{EEDL08}
{\sc W.~E and B.~Yu}, {\em The deep ritz method: {A} deep learning-based
  numerical algorithm for solving variational problems}, CoRR, abs/1710.00211
  (2017).

\bibitem{LW}
{\sc L.~Evans}, {\em Partial Differential Equations}, Graduate studies in
  mathematics, American Mathematical Society, 2010.

\bibitem{UQDL071}
{\sc G.~Fishman}, {\em Monte Carlo: concepts, algorithms, and applications},
  Springer Science \& Business Media, 2013.

\bibitem{oUQDL030}
{\sc M.~Frechet}, {\em On two new chapters in the theory of probability},
  Mathematics Magazine, 22 (1948), pp.~1--12.

\bibitem{DL}
{\sc I.~Goodfellow, Y.~Bengio, and A.~Courville}, {\em Deep Learning}, MIT
  Press, 2016.
\newblock \url{http://www.deeplearningbook.org}.

\bibitem{k002}
{\sc I.~Goodfellow, J.~Pouget-Abadie, M.~Mirza, B.~Xu, D.~Warde-Farley,
  S.~Ozair, A.~Courville, and Y.~Bengio}, {\em Generative adversarial nets},
  Advances in neural information processing systems, 27 (2014).

\bibitem{UQDL041}
{\sc W.~Huang, D.~Zhao, F.~Sun, H.~Liu, and E.~Chang}, {\em Scalable gaussian
  process regression using deep neural networks}, in Twenty-fourth
  international joint conference on artificial intelligence, 2015.

\bibitem{EEDL03}
{\sc S.~Karumuri, R.~Tripathy, I.~Bilionis, and J.~Panchal}, {\em
  Simulator-free solution of high-dimensional stochastic elliptic partial
  differential equations using deep neural networks}, Journal of Computational
  Physics, 404 (2020), p.~109120.

\bibitem{UQDL089}
{\sc D.~P. Kingma and J.~Ba}, {\em Adam: A method for stochastic optimization},
  2017.

\bibitem{UQDL004}
{\sc I.~Lagaris, A.~Likas, and D.~Fotiadis}, {\em Artificial neural networks
  for solving ordinary and partial differential equations}, IEEE Transactions
  on Neural Networks, 9 (1998), p.~987–1000.

\bibitem{Gfm001}
{\sc C.~Ledig, L.~Theis, F.~Huszar, J.~Caballero, A.~Cunningham, A.~Acosta,
  A.~Aitken, A.~Tejani, J.~Totz, Z.~Wang, and W.~Shi}, {\em Photo-realistic
  single image super-resolution using a generative adversarial network}, in The
  IEEE Conference on Computer Vision and Pattern Recognition (CVPR), July 2017.

\bibitem{ke6004}
{\sc C.~{Li}, Z.~{Wang}, and H.~{Qi}}, {\em Fast-converging conditional
  generative adversarial networks for image synthesis}, in 2018 25th IEEE
  International Conference on Image Processing (ICIP), Oct 2018,
  pp.~2132--2136.

\bibitem{UQDL088}
{\sc M.~Raissi, P.~Perdikaris, and G.~Karniadakis}, {\em Physics-informed
  neural networks: A deep learning framework for solving forward and inverse
  problems involving nonlinear partial differential equations}, Journal of
  Computational Physics, 378 (2019), pp.~686--707.

\bibitem{UQDL075}
{\sc M.~Raissi, P.~Perdikaris, and G.~E. Karniadakis}, {\em Inferring solutions
  of differential equations using noisy multi-fidelity data}, Journal of
  Computational Physics, 335 (2017), pp.~736--746.

\bibitem{UQDL076}
\leavevmode\vrule height 2pt depth -1.6pt width 23pt, {\em Numerical gaussian
  processes for time-dependent and non-linear partial differential equations},
  2017.

\bibitem{FPDL06}
{\sc J.~Sirignano and K.~Spiliopoulos}, {\em Dgm: A deep learning algorithm for
  solving partial differential equations}, Journal of Computational Physics,
  375 (2018), p.~1339–1364.

\bibitem{oUQDL007}
{\sc A.~J. Smola and B.~Sch\"{o}kopf}, {\em Sparse greedy matrix approximation
  for machine learning}, in Proceedings of the Seventeenth International
  Conference on Machine Learning, ICML '00, San Francisco, CA, USA, 2000,
  Morgan Kaufmann Publishers Inc., p.~911–918.

\bibitem{UQDL055}
{\sc E.~Snelson and Z.~Ghahramani}, {\em Sparse gaussian processes using
  pseudo-inputs}, in Advances in Neural Information Processing Systems,
  Y.~Weiss, B.~Sch\"{o}lkopf, and J.~Platt, eds., vol.~18, MIT Press, 2006.

\bibitem{UQDL068}
{\sc P.~Vincent, H.~Larochelle, Y.~Bengio, and P.-A. Manzagol}, {\em Extracting
  and composing robust features with denoising autoencoders}, in Proceedings of
  the 25th International Conference on Machine Learning, ICML '08, New York,
  NY, USA, 2008, Association for Computing Machinery, p.~1096–1103.

\bibitem{UQDL077}
{\sc Z.~Wang, W.~Xing, R.~Kirby, and S.~Zhe}, {\em Physics regularized gaussian
  processes}, 2020.

\bibitem{RW}
{\sc C.~K. Williams and C.~E. Rasmussen}, {\em Gaussian processes for machine
  learning}, vol.~2, MIT press Cambridge, MA, 2006.

\bibitem{UQDL046}
{\sc A.~G. Wilson, Z.~Hu, R.~Salakhutdinov, and E.~P. Xing}, {\em Deep kernel
  learning}, in Proceedings of the 19th International Conference on Artificial
  Intelligence and Statistics, A.~Gretton and C.~C. Robert, eds., vol.~51 of
  Proceedings of Machine Learning Research, Cadiz, Spain, 09--11 May 2016,
  PMLR, pp.~370--378.

\bibitem{oUQDL006}
{\sc A.~G. Wilson and H.~Nickisch}, {\em Kernel interpolation for scalable
  structured gaussian processes {(KISS-GP)}}, CoRR, abs/1503.01057 (2015).

\bibitem{UQDL070}
{\sc D.~Xiu and G.~E. Karniadakis}, {\em The wiener--askey polynomial chaos for
  stochastic differential equations}, SIAM Journal on Scientific Computing, 24
  (2002), pp.~619--644.

\bibitem{UQDL060}
{\sc X.~Yang, D.~Barajas-Solano, G.~Tartakovsky, and A.~M. Tartakovsky}, {\em
  Physics-informed cokriging: A gaussian-process-regression-based multifidelity
  method for data-model convergence}, Journal of Computational Physics, 395
  (2019), pp.~410--431.

\bibitem{UQDL050}
{\sc X.~Yang, G.~Tartakovsky, and A.~Tartakovsky}, {\em Physics-informed
  kriging: A physics-informed gaussian process regression method for data-model
  convergence}, 2018.

\bibitem{oUQDL002}
{\sc Y.~Yang and P.~Perdikaris}, {\em Adversarial uncertainty quantification in
  physics-informed neural networks}, Journal of Computational Physics, 394
  (2019), pp.~136--152.

\bibitem{UQDL078}
{\sc Y.~Yuan, Q.~Wang, and X.~T. Yang}, {\em Modeling stochastic microscopic
  traffic behaviors: a physics regularized gaussian process approach}, 2020.

\bibitem{UQDL079}
{\sc Y.~Yuan, Z.~Zhang, and X.~T. Yang}, {\em Highway traffic state estimation
  using physics regularized gaussian process: Discretized formulation}, 2020.

\bibitem{UQDL061}
{\sc Y.~Yuan, Z.~Zhang, X.~T. Yang, and S.~Zhe}, {\em Macroscopic traffic flow
  modeling with physics regularized gaussian process: A new insight into
  machine learning applications in transportation}, Transportation Research
  Part B: Methodological, 146 (2021), pp.~88--110.

\bibitem{UQDL009}
{\sc Y.~Zhu and N.~Zabaras}, {\em Bayesian deep convolutional encoder–decoder
  networks for surrogate modeling and uncertainty quantification}, Journal of
  Computational Physics, 366 (2018), p.~415–447.

\bibitem{UQDL002}
{\sc Y.~Zhu, N.~Zabaras, P.-S. Koutsourelakis, and P.~Perdikaris}, {\em
  Physics-constrained deep learning for high-dimensional surrogate modeling and
  uncertainty quantification without labeled data}, Journal of Computational
  Physics, 394 (2019), pp.~56--81.

\bibitem{oUQDL012}
{\sc M.~A. Álvarez, D.~Luengo, and N.~D. Lawrence}, {\em Linear latent force
  models using gaussian processes}, IEEE Transactions on Pattern Analysis and
  Machine Intelligence, 35 (2013), pp.~2693--2705.

\end{thebibliography}

\appendix
\section{Appendix: Implementation details}
\label{sec:App}

\begin{figure}
  \centering
  \includegraphics[scale=0.5]{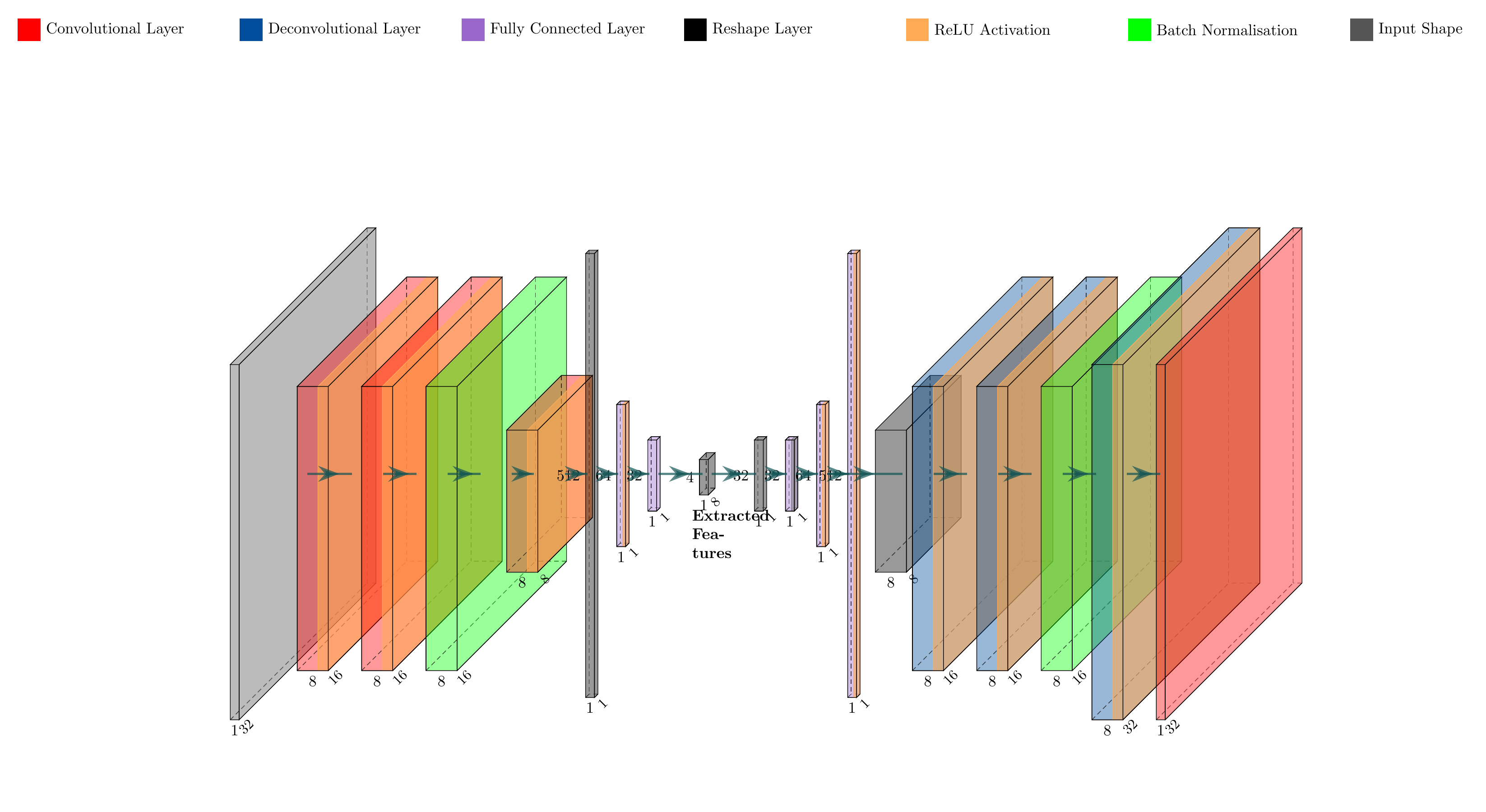}
  \caption{Network architecture in GRF experiment.}
  \label{fig:net}
\end{figure}

In this section we present the technical details regarding the neural network used in this work. Figure \ref{fig:net} shows the network structure used in the GRF experiment (section \ref{GRF})\footnote[2]{The LaTeX code template for plotting the neural network structure figure is from \url{https://github.com/HarisIqbal88/PlotNeuralNet}. Copyright (c) 2018 HarisIqbal88}. The feature in the middle "bottleneck" part is used in the covariance, while the final output is for calculating the loss at equation \ref{DAE_loss}. The network is a simple sequential model without shortcuts, and the number of layers is also small, thus will not consumes much resources. The batch size is set to be 96. To split it into 2 part as described in section \ref{subsec:SIP}, we simply divide each batch evenly, i.e. 48 data points in $B_k$ and another 48 in $B_{uk}$. In our numerical experiments, we simply set the hyper-parameters $\beta$ and $\gamma$ to be 1. All the experiments run on an NVIDIA RTX 2080 graphic card. We use Adam optimiser \cite{UQDL089}, with learning rate $0.0001$, and parameters $\beta_1$ and $\beta_2$ optimiser are set to be 0.9 and 0.999, respectively.

\begin{figure}
  \centering
  \includegraphics[scale=0.5]{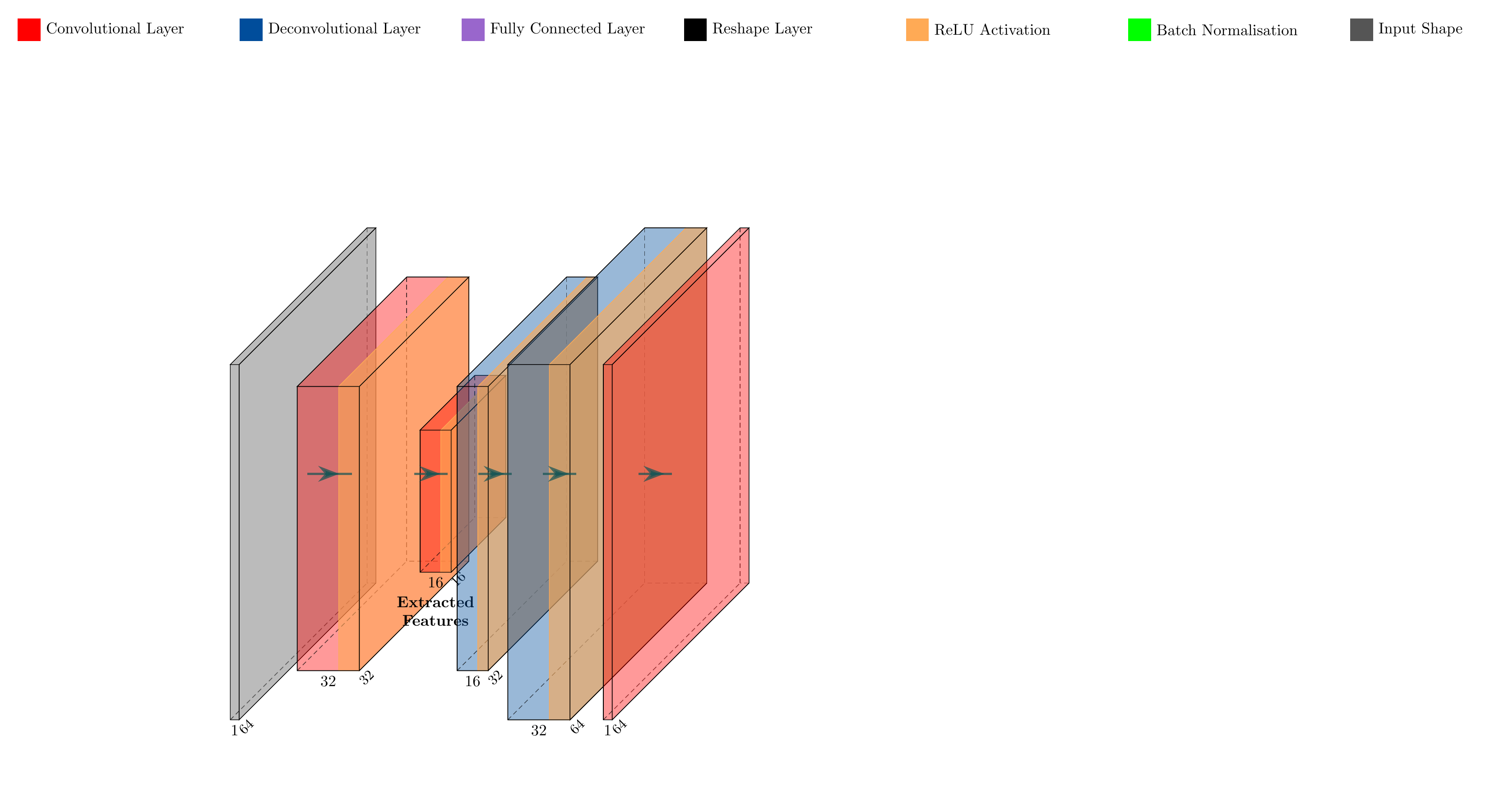}
  \caption{Network architecture in channelised field experiment.}
  \label{fig:net_64}
\end{figure}

In the channelised field experiment (section \ref{CF}), we use a similar structure. Figure \ref{fig:net_64} show the detailed structure. The batch size during training is the same as the previous experiment, i.e. 96, and is divided by half. The learning rate is $0.00001$. During testing, the batch size is set to be 768. This batch size means that we pick 768 data points from testing set one time. Other settings are the same as the previous experiment.

\end{document}